\documentclass{book}    
\usepackage{piers}  
\usepackage{multicol}
\usepackage{multirow}
\usepackage{comment}
\usepackage{amsmath,epsfig} 

\usepackage{amsmath,graphicx} 
\usepackage{subfig}
\usepackage{color,soul}
\usepackage{epstopdf}
\usepackage{tikz}
\usepackage{balance}
\usepackage{url}
\usepackage{multirow,hhline}
\usepackage{placeins}
\usepackage{wasysym}
\usepackage{amssymb}

\usepackage[nomessages]{fp}
\newcommand{\bx}{\mathbf{x}}
\newcommand{\by}{\mathbf{y}}
\newcommand{\bw}{\mathbf{w}}
\newcommand{\bz}{\mathbf{z}}

\newcommand{\bbb}{\mathbf{b}}

\newcommand{\ru}{\rule{0mm}{4mm}}

\begin{document}
\title{A CNN-Based Super-Resolution Technique for Active Fire Detection on Sentinel-2 Data}

\maketitle

\begin{authors}
{\bf  M. Gargiulo} , {\bf D. A. G. Dell'Aglio} , {\bf A. Iodice} , {\bf D. Riccio} , {\bf and G. Ruello} \\
\medskip
 University of Naples ``Federico II'', Italy\\
\end{authors}

\begin{paper}
\begin{piersabstract}
Remote Sensing applications can benefit from a relatively fine spatial resolution multispectral (MS) images and a high revisit frequency ensured by the twin satellites Sentinel-2.
Unfortunately, only four out of thirteen  bands  are provided at the highest resolution of 10 meters, and the others at 20 or 60 meters. For instance the Short-Wave Infrared (SWIR) bands, provided at 20 meters,  are very useful to detect active fires. Aiming to a more detailed Active Fire Detection (AFD) maps, we propose a super-resolution data fusion method based on Convolutional Neural Network (CNN) to move towards the 10-m spatial resolution the SWIR bands. The proposed CNN-based solution achieves better results than alternative methods in terms of some accuracy metrics. Moreover we test the super-resolved bands from an application point of view by monitoring active fire through classic indices. Advantages and limits of our proposed approach are validated on specific geographical area (the mount Vesuvius, close to Naples) that was damaged by widespread fires during the summer of 2017.
\end{piersabstract}

\psection{Introduction}
\label{sec:intro}
The remote sensing products are exploiting more and more in the earth monitoring because of the increasing number of satellites \cite{joshi2016review}. The European Space Agency has recently launched the twin satellites Sentinel-2 which can acquire global data for different applications such as risk management (floods, subsidence, landslide), 
land monitoring, water management, soil protection and so forth \cite{Drusch2012}. Sentinel-2 data are even useful in burnt area and active fire monitoring, using several algorithms \cite{verhegghen2016potential}. A plethora of these is essentially based on the threshold of spectral indices involving Near-Infrared (NIR) and Short-Wave Infrared (SWIR) bands \cite{cicala2018landsat,chuvieco97,Kant2000} that Sentinel-2 provides at spatial resolution of 10 m and 20 m, respectively. Therefore, it is common to resort to the 20-m resolution indices, by just downscaling the NIR band from 10 m to 20 m. However, following this approach spatial information from NIR band would be lost. An alternative approach to enhance the AFD method using the Sentinel-2 images is to produce the Active Fire Indices (AFIs) by upscaling the SWIR bands from 20 m to 10 m. Beyond the shadow of a doubt the main issue is the choice of the method to improve the spatial resolution of the SWIR bands. In general, Single Image Super Resolution (SISR) and Super Resolution Data Fusion (SRDF) methods are the two most popular ways to increase the spatial resolution of the images. The SISR methods do not use additional information from other sources, and they rely on spatial features of original image to increase its own resolution. On the other hand, SRDF methods (for instance, pan-sharpening) are based on the idea that the spatial information from other sources is useful to improve the spatial resolution of the original image \cite{Wald2002}. In order to produce the 10-m AFIs from Sentinel-2 bands with SRDF methods, the use of all the highest spatial resolution bands is not very benefiting because of their smoke-sensitivity. The major contribution is derived from the NIR which is the only band we consider in the SRDF approaches.\\
The rest of the paper is organized as follows. Section 2 describes the study area and the picked dataset. Section 3 gives more details about the methodology, focusing on the proposed CNN-based super-resolution method, hereafter $SRNN_{+}$, on the spectral fire indices and on the considered accuracy metrics. Section 4 summarizes experimental results, placing more attention on the super-resolved SWIR bands both in terms of visual inspection and numerical analysis while conclusions are drawn in Section 5.

\psection{Study Area and Dataset}

The area under investigation is located at the Vesuvius (in Fig. \ref{fig:fig0}), a volcano close to Naples,Italy. We are motivated by the choice of the study area since the presence of a natural park with a huge variety of flora and fauna considering its limited size \cite{mascolo2013vesuvius}. At the beginning of July 2017, hundreds of wildfires ignited and damaged across the Italy country, whose the most serious were at Vesuvius. In fact, fires had been interesting the Vesuvius area for several days and the situation quickly became more dangerous due to adverse climatic conditions (winds and dry weather) \cite{BOVIO2017e2537014}. The considered dataset is the Sentinel-2 Level-1C product acquired on 12th July 2017. As we can see in Fig. \ref{fig:fig0}, the area under investigation is mainly covered by heavy smoke (Fig. \ref{fig:fig0}-(b)) which reduces the usability of 10-m spectral information. 

\begin{figure}[!h]
\centering
\begin{tabular}{cc}
\includegraphics[width=30mm]{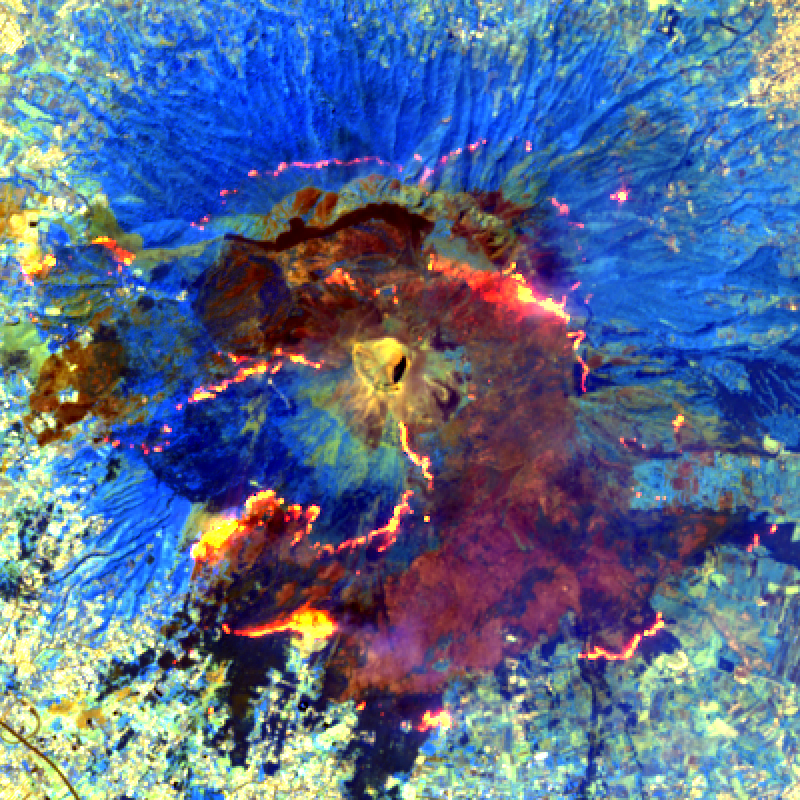} & \includegraphics[width=30mm]{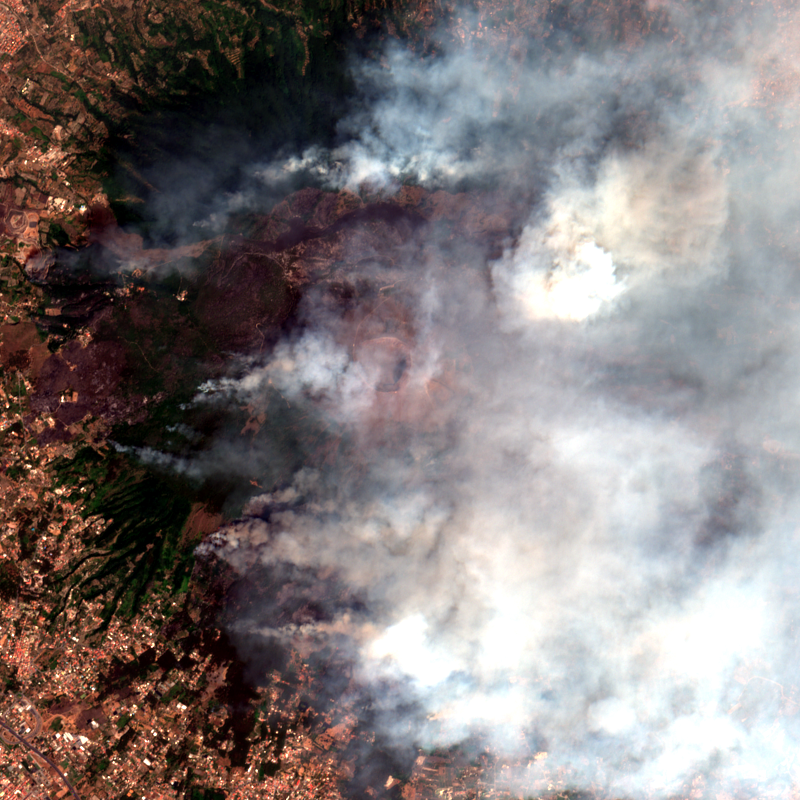}\\
(a) & (b) \\
\end{tabular}
\caption{ \small (a) false colour composite ($\rho_{12}$, $\rho_{11}$ and $\rho_{8}$ bands) and (b) RGB image of Vesuvius .}
\label{fig:fig0}
\end{figure}

\psection{Methodology}
\label{sec:method}
\psubsection{Proposed CNN-based Super-Resolution Fusion}

Our goal is to improve the spatial resolution of SWIR bands using a Convolutional Neural Network (CNN). CNNs have attracted an increasing interest in  many remote sensing applications, like object detection \cite{guo2018}, classification \cite{Krizhevsky2012}, pansharpening \cite{Scarpa2018}, and others, because of their capability to approximate complex non-linear functions, benefiting from the reduction in computation time obtained thanks to the GPU usage. On the downside the availability of a large amount of data is required for training. In this work we propose to use a relatively shallow architecture. This is composed by a cascade of $L=3$ convolutional layers. The first two are interleaved by Rectified Linear Unit (ReLU) activations that ensure fast convergence of the training process \cite{Krizhevsky2012}, and a linear activation function is considered in the last layer. The $l$-th ($1\leq l \leq 3$) convolutional layer, with $N$-band input $\bx^{(l)}$, yields an $M$-band output $\by^{(l)}$
\[     \by^{(l)} = \bw^{(l)} \ast \bx^{(l)} + \bbb^{(l)}. \]
In $l = 1$ case, the $\bx^{(l)} $ input is equal to the input, instead in $l = 3$ case, the  $\by^{(l)}$ is the output of the CNN. The tensor $\bw$ is a set of $M$ convolutional $N\times(K\times K)$ filters, where a $K\times K$ is the receptive field, while $\bbb$ is a $M$-vector bias. These parameters, $\Phi_l\triangleq\left(\bw^{(l)},\bbb^{(l)}\right)$, are refined during the training phase. Further information about the CNN  architecture can be found in \cite{Max2018}. In the supervised learning we need to generate a large amount of training samples, i.e. examples of inputs-target pairs. As reported in the pansharpening case \cite{Masi2016} the training samples have to ensue the Wald's protocol, that means to consider as inputs the downsampled PAN-MS pairs and taking as corresponding output the original MS. 

This approach has inspired our study where the highest spatial resolution bands play the role of the PAN and the MS is acted by SWIR. In our case we consider the training samples $\bx^{(1)}$ = ($ \mathbf{ \tilde{x}}$, $\mathbf{ \tilde{z} }$) as input to the network,  where $ \mathbf{ \tilde{x}}$ and $\mathbf{ \tilde{z} }$ are respectively the lower resolution version of the SWIR bands and of the 10-m bands provided by Sentinel-2, instead we consider the sharpened SWIR bands as output ($\by^{(3)}$ = $\bx $). Furthermore, the cost function and the learning optimization algorithm are required in the learning phase. In more details we use the L1-norm as cost function, in place of the L2-norm, to be more effective in error back-propagation when the computed errors are very low \cite{Scarpa2018}. Specifically, the loss is computed on the cost function by averaging over the training examples at each updating step of the learning process:
\[ L(\Phi^{(n)}) = {\rm E}\left[
			\left\| \bx  - \hat{\bx}(\Phi^{(n)}) \right\|_1
			\right]\]where $\bx$ represents the target and $\hat{\bx}$ the output of the CNN, dependent on the learnable parameters ($\Phi^{(n)}$) . Instead, in this work we use the  ADAM optimization method, an adaptive version of the Stochastic Gradient Descent (SGD), and it adapts the learning rates for each parameter of the CNN. This method requires very few tuning \cite{julian2017} and minimizes loss very speedily \cite{kingma2014}.

\psubsection{Spectral Fire Indices}

The proposed model is evaluated, from the application point of view, by monitoring active fires through the computation of three different spectral indices (in Fig. \ref{fig:fig00}), mainly used to this aim in literature \cite{Huang16,Schroder15,barducci2002infrared} because of their ease computing. The AFIs \cite{Huang16,Schroder15} are defined on Sentinel-2 data as follows:
\[\;\;\;
AFI_{1} = \frac{\rho_{12}}{\rho_{8}} \;\;\;\;\;\;\;\;\;\;
AFI_{2} = \frac{\rho_{11}}{\rho_{8}}
\;\;\;\;\;\;\;\;\;\;
AFI_{3} = \frac{\rho_{12}}{\rho_{11}} \;\;\; \]


where $\rho_{8}$ is the 10-m spatial resolution NIR band, centered at the wavelength of 0.834 $\mu$\textit{m}; while $\rho_{11}$ and $\rho_{12}$ are the 20-m SWIR bands, centered at 1.610 $\mu$\textit{m} and 2.190 $\mu$\textit{m}, respectively. All of these bands represent the radiance data at top of the atmosphere. The choice of these indices is based on their physical properties . Specifically, the conditions $AFI_{1} > 1$ and $AFI_{3} > 1$ often occur in active fire; while the condition $AFI_{2} < 1$ is verified near the fire fronts \cite{barducci2002infrared}.

\begin{figure}[!htbp]
\centering
\begin{tabular}{ccc}
\ru $AFI_{1}$ & $AFI_{2} $& $AFI_{3}$ \\ 
\includegraphics[width=30mm]{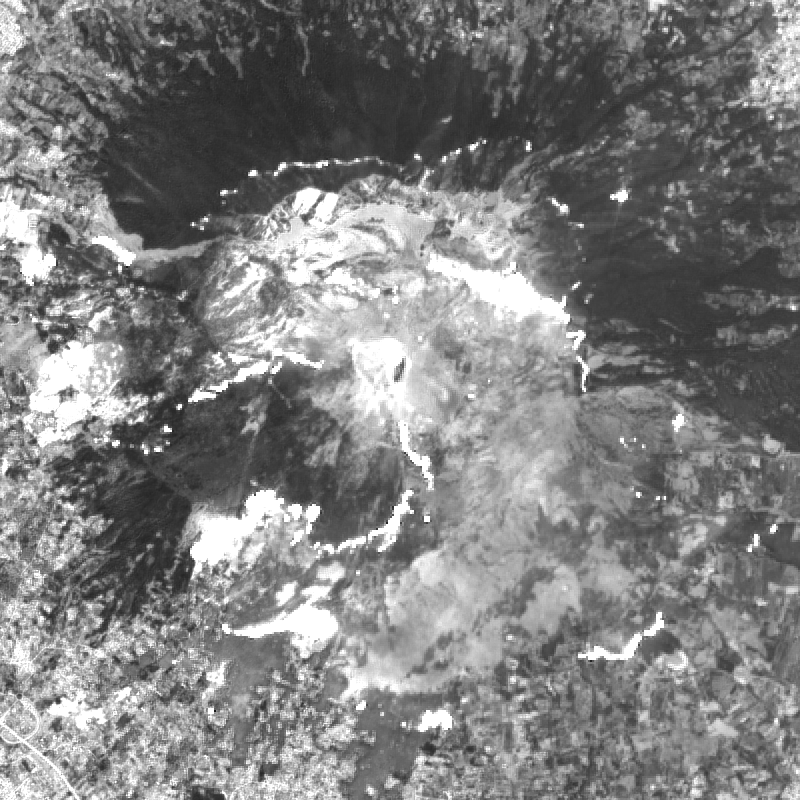} & \includegraphics[width=30mm]{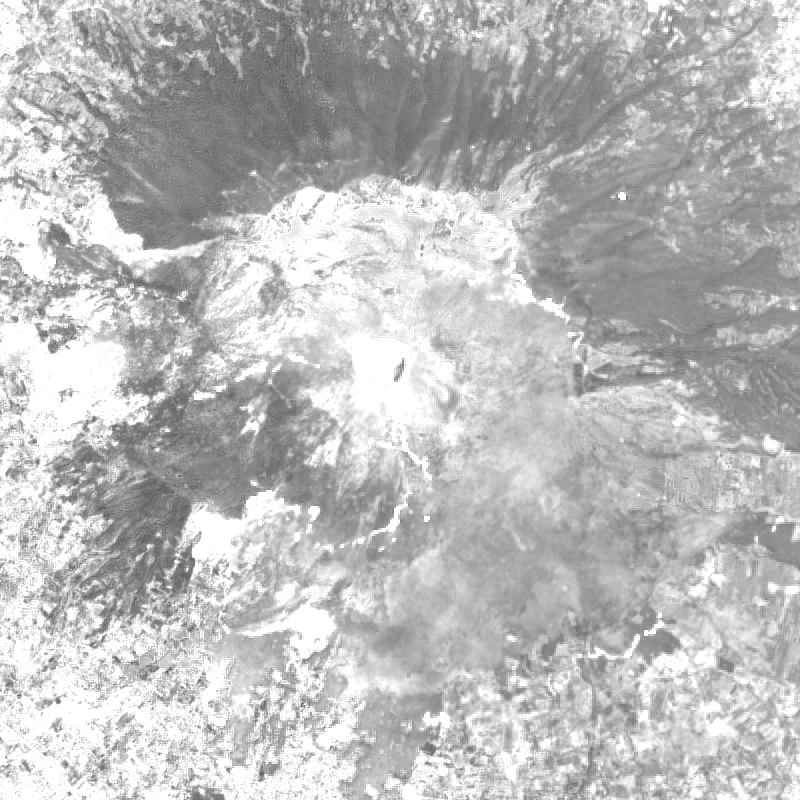}& \includegraphics[width=30mm]{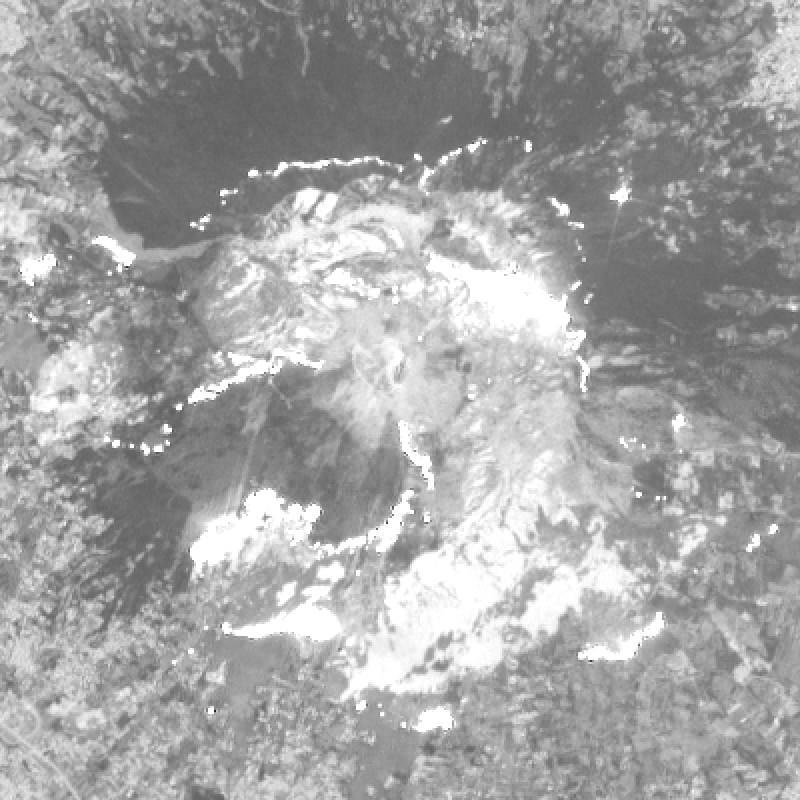} \\
\end{tabular}
\caption{\small Active Fire Indices related to Vesuvius.}
\label{fig:fig00}
\end{figure}

\psubsection{Results Accuracy Metrics}

To evaluate the performance when the target image is available (in our case, at 20-m spatial resolution), the proposed method is compared to alternative methods using four reference metrics, commonly used for pansharpening \cite{jagalingam2015review}:
\begin{itemize}
\setlength{\itemsep}{-2pt}
\item[-] Spectral Angular Mapper (SAM)  the spectral distortion between pixel of reference image and estimated one  \cite{alparone2007comparison};
\item[-] Universal Image Quality Index (UIQI, or Q-index), an image quality indicator introduced in \cite{Zhou2002};
\item[-] Relative Dimensionless Global Error (as known as ERGAS) which reduces to the root mean square error (RMSE) in case of single band  \cite{Wald2002};
\item[-] High-frequency Correlation Coefficient (HCC), the correlation coefficient between the
	high-pass components of two images \cite{Max2018}.
\end{itemize}
For a full resolution analysis we consider the active fire monitoring application, and all the methods compete with each others in terms of binary classification. To this end we need to define a ground truth on which computing the main classification metrics. In this context, such ground truth is performed with a differential multi-temporal approach, based on a thresholding of the difference between two cloudily-free realizations of Normalized Difference Vegetation Index (NDVI) in two different date (before and after the fire event). This ground truth (GT) is affected by noise (or small bright pixels) and so we have used a morphological operator (opening) to erase this undesired noisy effect.

 Thus, we compare this GT with the active fire maps obtained by thresholding the above-mentioned spectral indices. In our case, the AFIs use the super-resolved bands with the different considered approaches. In particular we consider different thresholds on each of AFIs to match the best detection of the active fires. In order to evaluate the quality of the obtained binary maps, we have considered some metrics, typically used in classification task: 
\begin{itemize}
\setlength{\itemsep}{-2pt}
\item[-] Precision (P) is the ratio between the correctly predicted positive observations and the total predicted positive ones;
\item[-] Recall (R) is the ratio between the correctly predicted positive observations to the all observations in actual class;
\item[-] Intersection over Union (IoU) is the ratio between the overlapping area and the union area. The intersection and the union are computed on the predicted positive observations and the positives from the GT.
\end{itemize}

It is worthwhile to remember that a high precision corresponds to a low false positive rate. In other words, higher the percentage of correctly predicted positive over the total predicted positive, higher precision.  Instead, a high recall corresponds to a low false negative rate, that means higher recall higher detection rate. 

\psection{ Results and Discussion}
\label{sec:results}

\psubsection{Training Phase}

Given the lack of sufficient available input-output samples in the present context, we start from a pre-trained CNN solution \cite{Max2018} to train the network's parameters $\Phi^n$. In \cite{Max2018} a super-resolution technique is considered for $ \rho_{11}$ band ($\bx =\rho_{11}$), and thus we extend to the $\rho_{12}$ band an equivalent solution ($\bx =\rho_{12}$). In particular, to create a pre-trained solution for this other band as well we use the identical dataset considered in \cite{Max2018}. After that, we fine-tune the weights of the CNN from Naples in two different dates, close to the target date (specifically June 27th and July 27th). This can be considered as a geographical fine-tuning, because we adapt the weights of the CNN on the geometric features of the study area. Then, we test this fine-tuned solution on the date under investigation (July 12th, 2017). Once left apart the target date for testing, 17$\times$17 patches for training are uniformly extracted from two above-mentioned date in the remaining segments. Overall, 10k patches are collected from the considered dates and randomly partitioned in 80\% for training phase and 20\% for validation phase. The 8k training patches are grouped in 32-size mini-batches for the implementation of the ADAM-based training. The fine-tuned solution is considered better than the solution from scratch, when a large amount of data for training phase is not available, or when the computing power is not sufficient \cite{tajbakhsh2016convolutional}. 
Eventually, we minimize the L1-norm cost function,  defined in the Methodology section, on the training examples using the ADAM learning algorithm. Thus, we set the ADAM default values $\eta$ = 0.002, $\beta_{1}$ = 0.9, and $\beta_{2}$ = 0.999, as reported in \cite{ruder2016}. In this specific case, the training phase requires 200 epochs ($32 \times 200$ weight update) performed in few minutes using GPU cards, while the test can be done in real-time.

\begin{figure}[!h]
\centering
\begin{tabular}{ccc}
\includegraphics[width=30mm]{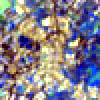} & \includegraphics[width=30mm]{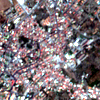}& \includegraphics[width=30mm]{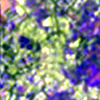} \\
Nearest Neighbour & $\bz$ & Bicubic \\[2mm]
\includegraphics[width=30mm]{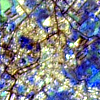} & \includegraphics[width=30mm]{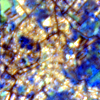}& \includegraphics[width=30mm]{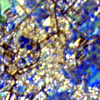} \\
 HPF & GS2-GLP &  $SRNN_{+}$ (Proposed)
\end{tabular}
\caption{\small Detail of the study area obtained by several super-resolution techniques and our proposal to underline the improvement in terms of spectral distortion. In the middle of the first row: $\bz$ is only composed by RGB bands.}
\label{fig:fig2}
\end{figure}

\psubsection{Comparison between Super-Resolution Proposal and SISR/SRDF techniques}

In this section, $SRNN_{+}$ is compared to a pre-trained CNN-based method (SRNN), three popular SRDFs adapted to the Sentinel-2/SWIR problem, including GS2-GLP \cite{vivone2015}, HPF \cite{Chavez1991} and PRACS \cite{vivone2015}, and even the SISRs, that are the Nearest Neighbour (NNI) and bicubic interpolation techniques. The numerical results obtained for the area of interest are reported in the left part of the Tab.~\ref{tab:SRtab}. In the results we consider an average on the SWIR bands. In the top part of the table, the $SRNN_{+}$ is compared to the SISR techniques, and the improvement is very remarkable in the HCC metric which deals with the fact that high frequency components are much affected by the super-resolution and are mostly localized on boundaries. Moving from the top to the bottom of the table, the proposed $SRNN_{+}$ method compares favourably against classical fusion methods, which take information from the additional input band $\rho_8$. In the penultimate row it is given the performance of the pre-trained SRNN model and even in this case the $SRNN_{+}$ performs slightly better results in terms of all the metrics at 20-m resolution. As it can be seen the additional fine-tuning, although having few training patches, provide a further gain. To conclude this section we show in Figg.~\ref{fig:fig1}-\ref{fig:fig2} some sample results (at 10-m without reference) which further confirm the effectiveness of the proposed method. 

\begin{figure}[!h]
\centering
\begin{tabular}{ccc}
\includegraphics[width=30mm]{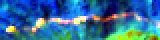} & \includegraphics[width=30mm]{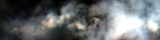}& \includegraphics[width=30mm]{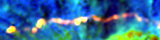} \\
Nearest Neighbour & $\bz$ & Bicubic \\[2mm]
\includegraphics[width=30mm]{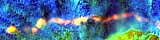} & \includegraphics[width=30mm]{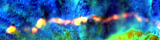}& \includegraphics[width=30mm]{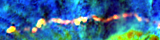} \\
 HPF  & GS2-GLP & $SRNN_{+}$ (Proposed)
\end{tabular}
\caption{ \small Detail of the area under investigation obtained by several super-resolution techniques and our proposal to underline the improvement in terms of spectral distortion. In the middle of the first row: $\bz$ is only composed by RGB bands that are affected by smoke presence (in the CNN input $\bz$ is also composed by $\rho_{8}$ band.}
\label{fig:fig1}
\end{figure}

\psubsection{Comparison between Different AFIs and Maps}

Once active fire ($AF$) is detected by considering the followed rules: $AFD = AFI_{k} >\alpha$, where $k \in \{1,2,3\}$, the performance is computed in terms of Precision, Recall and IoU and reported on the right-hand side of Tab.~\ref{tab:SRtab}. The numerical results confirmed the effectiveness of the proposal, and the Fig.\ref{fig:fig33}-\ref{fig:fig3} further confirm the superiority of the proposed method. As we can see in Tab. \ref{tab:SRtab}, the $SRNN_{+}$ have the best performance in terms of precision metric. In particular, its values are much greater than those relating to the classic techniques, demonstrating that it benefits from the joint information obtained from the visible bands. The low false alarm rate aspect is well visible in Fig.\ref{fig:fig3}, where an urban detail included in the study area is shown. The Fig.\ref{fig:fig3} only refers to $AFI_{2}$, but similar results are provided by the other indices analysed. On the other hand, both in terms of recall and IoU measures, the proposal have worse performance. We suppose this are mainly due to the ground truth used in validation, which probably over-estimates the areas interested by fires. In fact, as we can see in the central column of the figure \ref{fig:fig33}, the $SRNN_{+}$ AFIs better define these areas, resulting lighter and thinner then that ones obtained by other techniques. Furthermore we can observe from visual inspection that the boundaries are more evident considering $\rho_{12}$ and $\rho_{11}$ than $\rho_{12}$ and $\rho_{8}$. In general, even though this determines a low detection rate on the maps obtained by the $AFI_{3}$ with respect to $AFI_{1}$.

\begin{table}[!h]
\centering
\def\arraystretch{0.5}
\setlength{\tabcolsep}{3.2pt}
\footnotesize
\begin{tabular}{l|cccc||ccc}  
\hline
\ru Methods & SAM & Q-index & ERGAS & HCC &Precision&Recall&IoU\\
\ru  & (0) & (1) & (0) & (1)&(1)&(1)&(1)\\
\hline
\ru NNI & 0.001960 & 0.9182 & 9.353 & 0.1355 & 0.8329&0.5773&0.5309\\
\ru Bicubic & 0.001964 & 0.9515 & 7.155 & 0.471 &0.8387&0.5900&0.5471\\
\hline 
\ru HPF \cite{Chavez1991} & 0.064590 & 0.9405 & 8.150 & 0.2826 &0.7799&0.5991&0.5476\\
\ru PRACS \cite{vivone2015} & 0.001979 & 0.9535 & 7.057 & 0.5117 &0.7993&0.5987&0.5497\\
\ru GS2-GLP \cite{vivone2015} & 0.050190 & 0.9540 & 7.043 & 0.4694 &0.8008&\textbf{0.6131}&\textbf{0.5571}\\
\hline
\ru SRNN \cite{Max2018} & 0.001963 & 0.9688 & 5.943 & 0.6246 &0.8373&0.5649&0.5158\\
\ru $SRNN_{+}$ (Proposed) & \textbf{0.001956 }& \textbf{0.9743} & \textbf{5.425} & \textbf{0.6334}&\textbf{0.8414}&0.5642&0.5157\\
\hline
\end{tabular}
\caption{\small In the left part of the table: average results in terms of main metrics (at 20-m), typically used in pansharpening and super-resolution context. In the right part of the table: average results in terms of classification metrics. }
\label{tab:SRtab}
\end{table}

\begin{figure}[!h]
\centering
\begin{tabular}{ccc}
\multicolumn{2}{c}{\includegraphics[width=30mm]{PNG/B12_B08/lingua/B04_RGB.png}} & \includegraphics[width=30mm]{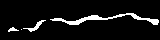} \\[2mm]
\multicolumn{2}{c}{} & (Ground-Truth) \\[2mm]
 \multirow{2}{*}{\includegraphics[width=30mm]{PNG/B12_B08/lingua/B12_Bicubic_RGB.png} }& \includegraphics[width=30mm]{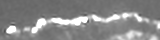}& \includegraphics[width=30mm]{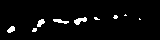} \\
 & \includegraphics[width=30mm]{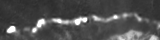}& \includegraphics[width=30mm]{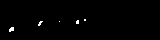} \\
& Bicubic & \\[2mm]
  \multirow{2}{*}{\includegraphics[width=30mm]{PNG/B12_B08/lingua/B12_GS2_GLP_RGB.png}} & \includegraphics[width=30mm]{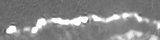}& \includegraphics[width=30mm]{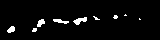} \\
 & \includegraphics[width=30mm]{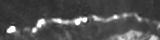}& \includegraphics[width=30mm]{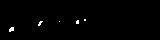} \\
& GS2-GLP & \\[2mm]
  \multirow{2}{*}{\includegraphics[width=30mm]{PNG/B12_B08/lingua/B12_FP_RGB.png}} & \includegraphics[width=30mm]{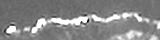}& \includegraphics[width=30mm]{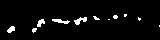} \\
 & \includegraphics[width=30mm]{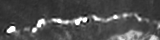}& \includegraphics[width=30mm]{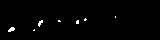} \\
& $SRNN_{+}$ (Proposed) & \\[2mm]
\end{tabular}
\caption{ \small In the first row the RGB image in which we can observe the presence of the smoke and the ground truth. Then, from the second row to the bottom: in the first column false-RGB, in the second $AFI_{1}$ and $AFI_{3}$, in the third the respective Maps. }
\label{fig:fig33}
\end{figure}

\begin{figure}[!h]
\centering
\begin{tabular}{ccc}
\multicolumn{2}{c}{\includegraphics[width=30mm]{PNG/B12_B08/false_alarm/B04_RGB.png}} & \includegraphics[width=30mm]{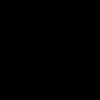} \\[2mm]
\multicolumn{2}{c}{} & (Ground-Truth) \\[2mm]
 \includegraphics[width=30mm]{PNG/B12_B08/false_alarm/B11_Bicubic_RGB.png} & \includegraphics[width=30mm]{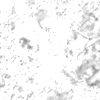}& \includegraphics[width=30mm]{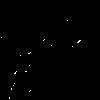} \\
& Bicubic & \\[2mm]
 \includegraphics[width=30mm]{PNG/B12_B08/false_alarm/B12_GS2_GLP_RGB.png} & \includegraphics[width=30mm]{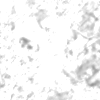}& \includegraphics[width=30mm]{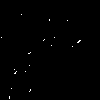} \\
& GS2-GLP & \\[2mm]
 \includegraphics[width=30mm]{PNG/B12_B08/false_alarm/B12_FP_RGB.png} & \includegraphics[width=30mm]{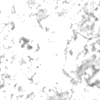}& \includegraphics[width=30mm]{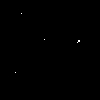} \\
& $SRNN_{+}$ (Proposed )& \\[2mm]
\end{tabular}
\caption{ \small In the first row the RGB image in which we can observe the absense of the smoke and the ground truth. Then, from the second row to the bottom: in the first column false-RGB, in the second $AFI_{2}$, and in the third the respective Map.}
\label{fig:fig3}
\end{figure}

\psection{Conclusion}

In this work we propose the $SRNN_{+}$ to further enhance the spatial resolution of the Sentinel-2 SWIR bands. For the specific goal (AFD) we fine-tune the weights of the CNN on the geographic study area and then we test the proposed approach both in terms of visual quality assessment and AFD capability. Eventually we show very promising results in terms of super-resolution metrics and even in AFD. The achieved results encourage us to exploit different architectural choices and/or learning strategies, and extending this approach to other remote sensing applications. 

\bibliographystyle{IEEEbib}
\bibliography{refs}
\end{paper}
\end{document}